\documentclass[]{interact}

\usepackage{epstopdf}
\usepackage[caption=false]{subfig}

\usepackage[numbers,sort&compress]{natbib}
\bibpunct[, ]{[}{]}{,}{n}{,}{,}
\renewcommand\bibfont{\fontsize{10}{12}\selectfont}
\makeatletter
\def\NAT@def@citea{\def\@citea{\NAT@separator}}
\makeatother

\theoremstyle{plain}

\theoremstyle{definition}

\theoremstyle{remark}

\usepackage{multirow}
\usepackage{booktabs}
\usepackage{bm}
\usepackage{url}
\usepackage{colortab, array, xcolor}
\usepackage{mathtools}
\usepackage{tabularx}
\usepackage{makecell}
\usepackage[whole]{bxcjkjatype} 
\usepackage{comment}
\usepackage{amssymb}
\usepackage{siunitx}
\usepackage{amsmath}

\def\it_d{ \mathrm{d} }
\def\tild_r{ \tilde{\bmr} }

\def\dfrac{ \displaystyle\frac}

\begin{document}

\articletype{ARTICLE TEMPLATE}

\title{Adaptive Undulatory Locomotion of Snake-like Robots in Dynamic Viscous Environments via Deep Reinforcement Learning}

\author{
\name{Tsuyoshi Kimoto\textsuperscript{a}\thanks{CONTACT Tsuyoshi Kimoto. Email: su23239o@st.omu.ac.jp}, Akio Yamano\textsuperscript{a}, Kohei Honda\textsuperscript{b} and Takashi Iwasa\textsuperscript{a}}
\affil{\textsuperscript{a}Osaka Metropolitan University, 1-1, Gakuen-cho, Naka-ku, Sakai, Osaka, Japan; \textsuperscript{b}Nagoya University, Chikusa-ku, Nagoya, Aichi, Japan}
}

\maketitle

\begin{abstract}
This paper demonstrates how deep reinforcement learning (DRL) enables adaptive locomotion of snake-like robots in dynamically changing viscous environments.
Traditional control methods exhibit inherent performance limitations in such environments.
The fundamental challenge of this task is the necessity to adapt to fluid properties that are unobservable with local onboard sensors.
To overcome this, we formulate the problem as a partially observable Markov decision process (POMDP) and solve it using an asymmetric actor-critic framework.
In this approach, a teacher policy trained using privileged information available only in the physics simulator distills its knowledge into a student policy that relies on proprioceptive sensor information.
Simulation results across a wide range of dynamic viscosity changes ($10^{-7}$ to $10^{-2}\mathrm{ m}^2\mathrm{/s}$) reveal that the DRL agent autonomously acquires non-sinusoidal adaptive gaits. 
These gaits improve propulsion velocity and transport efficiency, breaking the inherent limits of conventional sinusoidal and kinematic control.
The findings establish that implicit environment inference via privileged information distillation is an effective approach to bypass the constraints of traditional models under unpredictable fluid dynamics.
\end{abstract}


\section{Introduction}
Snake-like robots are promising for navigating heterogeneous environments like flooded infrastructure~\cite{Ji2023, Nivethika2022} and muddy disaster sites~\cite{Kamegawa2020} where fluid properties vary spatially and temporally.
Their slender, multi-jointed bodies enable effective navigation through narrow spaces via undulatory locomotion.
However, undulatory swimming performance depends on kinematic viscosity fluctuating with temperature changes and impurities.
Relying on fixed, pre-designed gaits leads to degradation in propulsion efficiency under such environmental uncertainty~\cite{Iwasaki2014, Yamano2023, Kimoto2025}.
This study aims to enable snake-like robot adaptation of undulatory locomotion to unknown and time-varying fluid viscosity.

Environment-adaptive undulatory locomotion is challenging because accurately modeling the fluid dynamics is difficult, and directly measuring environmental changes onboard is impractical.
Model-based control requires solving complex fluid dynamics such as the Navier-Stokes equations, which is computationally infeasible for real-time motion generation~\cite{Abouhussein2023}.
Viscometers are unsuitable for compact and waterproof snake-like robots, and pre-designed gait switching cannot cover continuously changing parameters in unknown environments.
Indicators of performance degradation such as propulsion velocity cannot be directly measured by the robot and must be estimated.
However, estimating velocity requires estimating fluid resistance, which not only has issues of model complexity and computational complexity but also a circular problem where viscosity is needed for calculating the resistance.
Real-time locomotion control under partially observable conditions remains the core challenge.

To overcome these limitations, this study demonstrates how deep reinforcement learning (DRL) can break the performance barriers of traditional control methods based on fixed, pre-designed gaits for snake-like robots.
Specifically, we formulate this challenging fluid robotics task as a partially observable Markov decision process (POMDP)~\cite{Kaelbling1998}.
In this formulation, the robot receives only proprioceptive sensor data, while the kinematic viscosity and translational velocity remain hidden.
We apply an asymmetric actor-critic paradigm, where a teacher policy trained with privileged information (viscosity and robot velocity) distills its knowledge into a student policy relying solely on onboard proprioceptive sensing~\cite{Chen2020}.
This framework successfully establishes a mechanism to implicitly infer fluid resistance and generate optimal joint target angles in real time without external measurement.

We validate this approach through extensive physics simulations and cross-environment evaluations targeting a wide range of kinematic viscosities.
The results demonstrate that the DRL policy autonomously acquires dynamic, non-sinusoidal waveforms that consistently surpass the optimal performance limits of conventional sinusoidal control.
Furthermore, comparing it with a standard symmetric DRL approach highlights the inadequacy of standard methods, demonstrating that the proposed privileged information distillation is essential for acquiring an effective swimming gait in this task.
These findings highlight that DRL, when appropriately formulated, provides a decisive advantage over traditional methods for robust real-time gait adaptation in heterogeneous fluid environments.

\section{Related work}
Research on propulsion control of snake-like robots is diverse.
Various types of snake-like robots and gaits have been proposed for different scenarios and tasks.
We focus on snake-like robots that do not have propulsion mechanisms such as wheels~\cite{Fukuoka2023} or screws~\cite{Vaquero2024}, which are structurally closer to biological snakes and have high off-road capabilities, assuming they are in fluids.

\subsection{Control of snake-like robots}
Hirose et al. proposed the serpenoid curve as a fundamental propulsion method for snake-like robots~\cite{Hirose2009}.
This curve changes curvature sinusoidally along the body length.
Since the tangent angle, which is the integral of curvature, also becomes sinusoidal, in discrete robots, joint angles are sometimes controlled as sinusoidal waves~\cite{Yamano2021,Yamano2023,Kimoto2025}.
This method is mathematically simple and allows for easy design of efficient propulsion motions.
However, these methods are constrained to specific periodic waveforms that are predefined, making it difficult to flexibly change motion according to changes in the fluid.

CPG (Central Pattern Generator), which mimics the nervous system of living organisms, has also been used~\cite{Manzoor2019}.
CPG can maintain a stable rhythm against external disturbances.
However, when environmental changes are significant, it requires complicated parameter tuning.
Moreover, the generated motion patterns are constrained to waveforms as specific periodic functions.
Similar to control using sinusoidal torque or serpenoid curves, it is difficult to flexibly change motion according to changes in the fluid.

Recently, DRL has been proposed to optimize policies through direct environment interaction~\cite{Baysal2025}.
Some studies utilize DRL to adjust complex CPG parameters , aiming to achieve both gait stability and adaptability~\cite{Liu2023,Liu2025}.
Nevertheless, these methods still depend on periodic, stereotyped CPG waveforms, limiting their flexibility in unknown environments.
This study instead adopts an end-to-end network that directly generates target joint angles, allowing for free waveform generation unconstrained by conventional sinusoidal frameworks.

\subsection{Adaptation to varying fluid viscosity}
The swimming performance of snake-like robots performing undulatory locomotion in fluids depends on the kinematic viscosity and density of the fluid.
In the biological world, nematodes~\cite{Gjorgjieva2014}, leeches~\cite{Iwasaki2014} and lungfish~\cite{Horner2008} change their frequency and wavelength of motion according to viscosity, maintaining high propulsion efficiency under various conditions.
As an engineering approach, Yamano et al. have conducted analysis using multi-objective optimization (MOO)~\cite{Yamano2023}.
This method can quantitatively show the optimal motion parameters for specific environments as a Pareto front.
However, performing real-time re-optimization against unknown environmental changes is difficult from the perspective of computational load.

There are also studies that have implemented the method of changing the motion pattern of nematodes in snake-like robots~\cite{Yamano2018}.
This method has the advantage of being able to automatically adjust the gait based on interaction with the environment.
However, this method requires prior environmental knowledge for parameter adjustment according to the fluid properties, so it cannot adapt to unknown environmental changes, and the range of motion changes is limited~\cite{Kimoto2021}.
Methods based on simulation depend on the accuracy of the fluid model, but in this approach, once a policy is learned, it can adapt in real time to unknown viscosity values.

In our DRL-based end-to-end approach, the advantage is that it implicitly identifies environmental characteristics from only internal sensor information without any direct measurement by external sensors.
By leveraging privileged information on the physics simulator for knowledge distillation, the robot can estimate the surrounding fluid resistance from its past motion-response history and generate optimal joint target angles in real time.
This enables a control system that can robustly adapt to a wide range of dynamic viscosity changes without requiring tuning for each environment.

\section{Undulatory locomotion of snake-like robots with dynamic fluid viscosity}
\subsection{Task definition and requirements}
\begin{figure}[!t]
        \centering
        \includegraphics[width=0.7\linewidth]{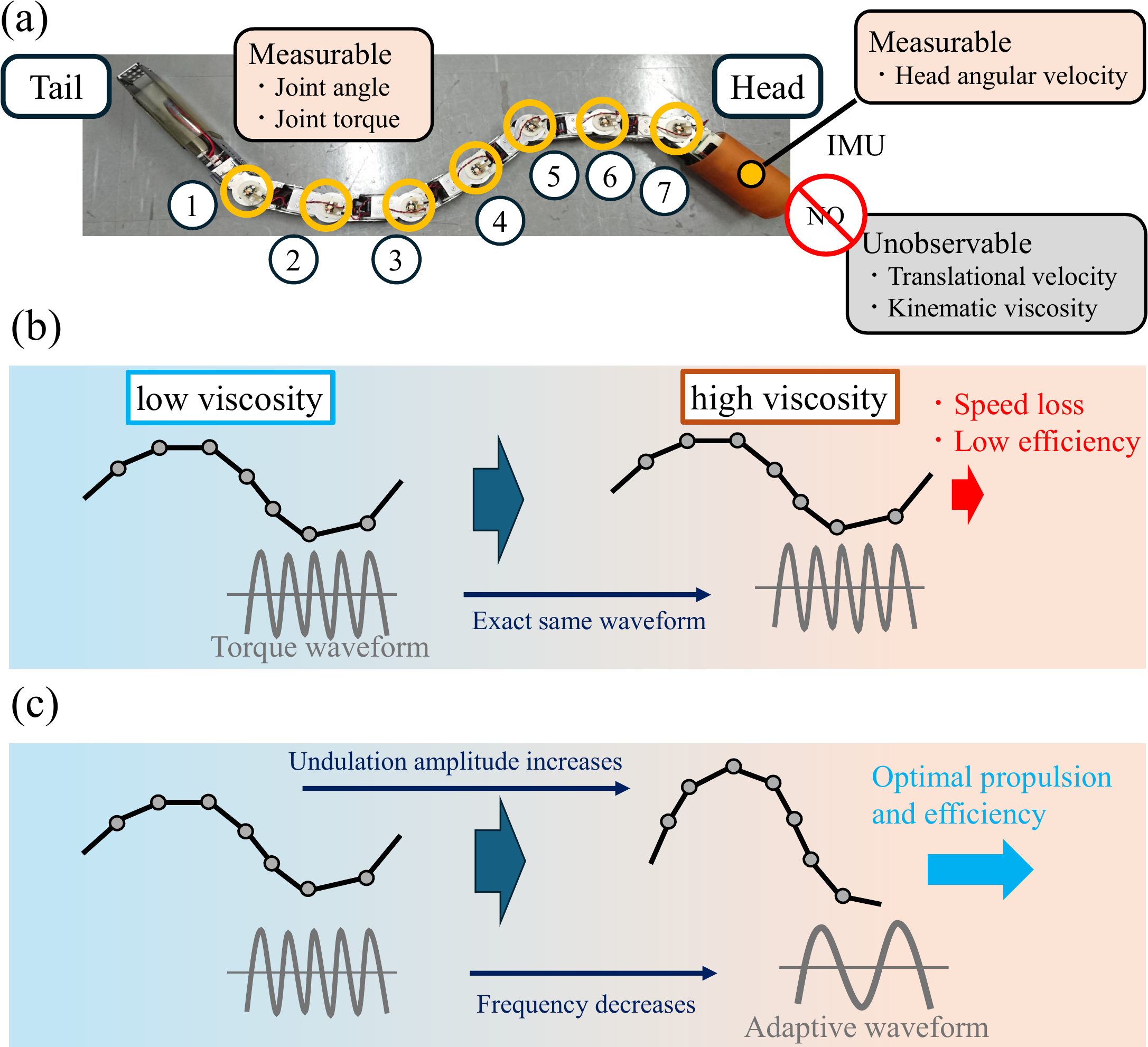}
        \caption[]{
                Conceptual overview of the environment-adaptive undulatory locomotion task.
(a) Appearance and sensor configuration of the snake-like robot. While joint states and head angular velocity are measurable via onboard sensors, the translational velocity and fluid kinematic viscosity remain unobservable.
(b) Performance degradation with a fixed gait. Maintaining a constant undulatory pattern as the environment transitions from low to high viscosity leads to significant speed loss and reduced efficiency.
(c) Proposed adaptive locomotion via the asymmetric actor-critic framework. The robot autonomously modifies its gait—decreasing frequency and increasing amplitude—to maintain optimal propulsion and efficiency without external sensors.
        }\label{fig:tisser_figure}
\end{figure}

We define the task of adaptive undulatory locomotion in unknown viscous environments.
Figure~\ref{fig:tisser_figure} illustrates the conceptual overview of the task targeted.
The swimming performance of snake-like robots depends on the kinematic viscosity of the fluid, which can fluctuate spatially and temporally in real environments such as disaster sites due to temperature changes and impurities.
As shown in Figure~\ref{fig:tisser_figure}(b), maintaining a fixed motion pattern against environmental changes leads to degradation in propulsion speed and efficiency as fluid resistance increases.
The objective of this study is to construct a control system that autonomously modifies the gait according to the resistance characteristics of the environment, as illustrated in Figure~\ref{fig:tisser_figure}(c), to maintain propulsion performance.

The appearance of the snake-like robot targeted is shown in Figure~\ref{fig:tisser_figure}(a), and its specifications are listed in Table~\ref{tab:Specification_of_the_snake-like_robot}.
The robot is a multi-joint mechanism with a total length of $\SI{1490}{mm}$ and a total mass of $\SI{5.50}{kg}$.
Of the 13 joints, 7 servomotors are used to generate undulatory locomotion in the horizontal plane.
The snake-like robot is covered with a rubber tube and nylon cloth for waterproofing, and it is equipped with a battery for autonomous swimming inside~\cite{Kimoto2023,Yamano2023}.

\begin{table}[!t]
        \tbl{Specification of the snake-like robot~\cite{Kimoto2023,Yamano2023}.}
        {\centering
        \begin{tabular}{lll}
                \hline
                \multirow{4}{*}{Length}& Total & $\SI{1490}{mm}$ \\
                & Head & $\SI{280}{mm}$ \\
                & Middle & $\SI{148}{mm} \times 6$ \\
                & Tail & $\SI{322}{mm}$ \\
                \hline
                \multirow{5}{*}{Mass}& Total & $\SI{5.50}{kg}$ \\
                & Head & $\SI{991}{g}$ \\
                & Middle & $\SI{462}{g} \times 6$ \\
                & Tail & $\SI{1194}{g}$ \\
                & Waterproof cover & $\SI{542}{g}$\\
                \hline
                Number of joints & & 13 (on the ground) \\
                & & 7 (in a fluid)\\
                \hline
        \end{tabular}}
        \label{tab:Specification_of_the_snake-like_robot}
\end{table}

\subsection{Formulation using POMDP}
We model the control problem of undulatory locomotion in unknown dynamic viscosity environments as a deterministic POMDP~\cite{Kaelbling1998}.
A POMDP is defined by the tuple $(\mathcal{S}, \mathcal{A}, T, R, \Omega, \mathcal{O}, \gamma)$, where $\mathcal{S}$ is the set of states, $\mathcal{A}$ is the set of actions, $T$ is the state transition function, $R$ is the reward function, $\Omega$ is the set of observations, $\mathcal{O}$ is the observation function, and $\gamma$ is the discount factor.
The state $\bm{s}_{n} \in \mathcal{S}$ consists of a combination of the robot's own state and the surrounding environmental conditions, where $n \in \mathbb{N}$ is the time step.
Specifically, it includes joint angles and angular velocities, as well as actuator torques.
Furthermore, it includes the kinematic viscosity $\nu$ of the fluid and the translational velocity $\bm{v}_n$ of the center of mass, which are important hidden variables in this task.
Joint angles and angular velocities can be observed by sensors, but the exact kinematic viscosity and translational velocity cannot be directly obtained from onboard sensors of the robot, and are treated as partially observable states.
The action $\bm{a}_n \in \mathcal{A}$ is a set of target angles for the 7 joints that generate undulatory locomotion in the horizontal plane.
Each action is defined as a 7-dimensional vector of continuous values $\bm{a}_n \in [-1, 1]^7$.


The state transition function $T: \mathcal{S} \times \mathcal{A} \mapsto \mathcal{S}$ describes the changes in the physical world when action $\bm{a}_n$ is applied.
In the transition process, the applied torques and the surrounding fluid resistance interact, resulting in the new posture and velocity of the robot at the next time step $n+1$ through fluid-structure interaction.
The physics simulation on the simulator serves as the role of this transition function.
The agent cannot directly know the state $\bm{s}_n$ but receives only observations $\bm{o}_n \in \Omega$ based on the observation model $\bm{o}_n = \mathcal{O}(\bm{s}_n)$.
The reward function $R: \mathcal{S} \times \mathcal{A} \mapsto \mathbb{R}$ quantifies the goodness of propulsion performance at each time step.
The specific composition of the observation space and the design of the reward function will be described in detail in Section 4.

\subsection{Joint dynamics and actuator Model}
Before detailing the specific control strategies, we formulate the physical dynamics of the robot's joints.
The command torques $u_{\mathrm{in},n}$ for each joint at time step $n$ are first generated by a PD controller:
\begin{equation}
u_{\mathrm{in},n} = K_p(a_n - q_n) - K_d \dot{q}_n
\label{eq:pd_control}
\end{equation}
where $K_p$ and $K_d$ are the proportional and derivative gains, respectively, and $a_n$ is the action defined in Section 3.2.
$q_n$ and $\dot{q}_n$ are the current joint angle and angular velocity.
Each joint generates the final output torque $u_n$ through a virtual torsion spring and a damper, which models the passive resistance of the waterproof cover:
\begin{equation}
u_n = u_{\mathrm{in},n} - k_\mathrm{body}q_n - c_\mathrm{body}\dot{q}_n
\label{eq:torque_generation}
\end{equation}
where $k_\mathrm{body}$ and $c_\mathrm{body}$ are the stiffness of the torsion spring and the viscosity coefficient of the damper, respectively.
This mechanism allows the robot to utilize passive restoring forces during locomotion.
The specific parameter values used for simulation will be detailed in Section 5.2.3.

\subsection{Existing strategy for the dynamic fluid viscosity}
To evaluate the performance of the proposed DRL framework, we establish two representative traditional optimization methods as baselines.
Because there are no direct competitor learning algorithms for this specific real-time fluid adaptation task, establishing the upper bounds of traditional control paradigms provides the most rigorous and sufficient benchmark.
The first is a Single-Objective Optimization (SOO) applied to a kinematic sea snake model, which evaluates comprehensive swimming performance using a unified scalar objective.
The second is a MOO applied to a sinusoidal torque control model, which explores the trade-off between propulsion speed and energy efficiency.
By comparing our learning-based approach against these two distinct traditional strategies, we can quantitatively verify its superiority.

\subsubsection{Single-objective optimization of a kinematic sea snake model}
As a standard swimming gait, we utilize a kinematic model inspired by sea snakes~\cite{Graham1987, Huang2022}.
The gait is described by a traveling wave where the amplitude increases from the head to the tail.
The target joint angle $a_i(t)$ for the $i$-th joint is defined as follows:
\begin{equation}
a_i(t) = \frac{i}{N_\mathrm{servo}} A_a \sin(2\pi f_\mathrm{cmd} t + (N_\mathrm{servo} - i)\psi_a)
\end{equation}
where $i \in \{1, \dots, N_\mathrm{servo}\}$ is the joint index, $f_\mathrm{cmd}$ is the frequency, $A_a$ is the angle amplitude, and $\psi_a$ is the phase difference between adjacent joints. 
This predefined continuous function $a_i(t)$ directly determines the action $a_n$ inputted to the PD controller in Eq.~\eqref{eq:pd_control}.

To derive the optimal kinematic parameter set $p_\mathrm{soo} \coloneqq [A_a, f_\mathrm{cmd}, \psi_a]$, we formulate a SOO to maximize a scalar objective function $J(p_\mathrm{soo})$:
\begin{equation}
\max_{p_\mathrm{soo} \in \mathcal{P}_\mathrm{soo}} J(p_{soo})
\end{equation}
where $\mathcal{P}_\mathrm{soo} \subset \mathbb{R}^3$ represents the feasible parameter space bounded by the kinematic constraints of the robot. 

To ensure a fair comparison with the proposed learning-based approach, the objective function $J(p_\mathrm{soo})$ is designed to match the evaluation criteria of the DRL agent.
Specifically, it is defined as the cumulative sum of the reward function $r_n$ over one episode:
\begin{equation}
J(p_\mathrm{soo}) \coloneqq \sum_{n=0}^{N_\mathrm{step}} r_n
\end{equation}
where $N_\mathrm{step}$ is the total number of control steps.
The components of $r_n$, which balance forward velocity, energy consumption, and movement penalties, will be detailed in Section 4.3.
This optimization yields a single optimal parameter set that maximizes the comprehensive swimming performance under predefined kinematic waves.

To solve this optimization problem efficiently, we utilize the Tree-structured Parzen Estimator (TPE) algorithm implemented via the Optuna framework~\cite{Akiba2019}.
Because evaluating the objective function $J(p_\mathrm{soo})$ requires executing a full simulation episode to calculate the cumulative reward, this sample-efficient Bayesian optimization approach is suitable.
The search space for the kinematic parameters is defined as follows: $A_a \in [0.1, \pi/3]$ rad, $f_\mathrm{cmd} \in [0.1, 0.8]$ Hz, and $\psi_a \in [0.1, 3\pi/7]$ rad.
The bounds for the amplitude $A_a$ and the frequency $f_\mathrm{cmd}$ are determined by the hardware structural constraints and the maximum performance limits of the actuators, respectively.
The range for the phase difference $\psi_a$ is set to maintain consistency with the optimal regions identified in previous multi-objective optimization study~\cite{Yamano2023}.
Furthermore, this range encompasses the kinematic characteristics observed in biological undulatory swimmers, in which the wavelength associated with the undulatory motions scales with the body length~\cite{Gazzola2014}, thereby providing a sufficient functional margin for the optimization process.

\subsubsection{Multi-objective optimization of sinusoidal torque control}
As a baseline for torque-based propulsion, we use a method based on sinusoidal control.
The command torque $u_{\mathrm{in},t}$ for the joints is defined as follows:
\begin{equation}
u_{\mathrm{in},t} \coloneqq 
\begin{bmatrix}
A_\mathrm{in} \sin(2\pi f_{cmd} t + (1-1)\psi_\mathrm{in}) \\
A_\mathrm{in} \sin(2\pi f_\mathrm{cmd} t + (2-1)\psi_\mathrm{in}) \\
\vdots \\
A_\mathrm{in} \sin(2\pi f_\mathrm{cmd} t + (N_\mathrm{servo}-1)\psi_\mathrm{in})
\end{bmatrix}
\end{equation}
where $A_\mathrm{in}$ is the amplitude and $\psi_\mathrm{in}$ is the phase difference between adjacent joints.
For this torque-based method, the continuous command torque $u_{\mathrm{in},t}$ is directly provided as an input to the joint dynamics in Eq.~\eqref{eq:torque_generation}, bypassing the PD controller.

By using MOO, we can derive the optimal parameter set for specific known environments \cite{Yamano2023, Kimoto2025}.
MOO is formulated as follows, where the design variable vector $p_\mathrm{moo} \coloneqq [A_\mathrm{in}, f_\mathrm{cmd}, \psi_\mathrm{in}]$ is the optimization target:
\begin{equation}
\min_{p_\mathrm{moo} \in \mathcal{P}_\mathrm{moo}} J_1(p_\mathrm{moo}), J_2(p_\mathrm{moo})
\end{equation}
where $\mathcal{P}_\mathrm{moo} \subset \mathbb{R}^3$ represents the feasible parameter space bounded by the physical constraints of the actuators. 

Unlike SOO, which aggregates factors into a single score, MOO evaluates the trade-off between conflicting physical metrics. 
The two objective functions to be minimized are the negative forward propulsion velocity and the actuator power consumption:
\begin{equation}
J_1(p_\mathrm{moo}) = -v, \quad J_2(p_\mathrm{moo}) = P
\end{equation}
where $v$ and $P$ are evaluation metrics defined later in Section 5.3.
Because this MOO is conducted through exhaustive evaluation up to the physical limits of the actuators, the resulting Pareto front represents the upper bound of propulsion performance for the fixed sinusoidal torque control paradigm.
The search space for this exhaustive evaluation is defined as follows: $A_\mathrm{in} \in [0.1, 4.0]$ Nm, $f_\mathrm{cmd} \in [0.1, 0.8]$ Hz, and $\psi_\mathrm{in} \in [0.1, 3\pi/7]$ rad.
Note that this amplitude bound applies to the command torque $u_{\mathrm{in},t}$; the final output torque $u_n$ may exceed this limit during locomotion due to the integration of passive restoring forces as formulated in Eq.~\eqref{eq:torque_generation}.
Furthermore, because this torque-based approach does not impose strict limits on the resulting joint angles, its feasible parameter space $\mathcal{P}_\mathrm{moo}$ is not strictly identical to the kinematically bounded space $\mathcal{P}_\mathrm{soo}$.

\section{DRL policy training with physics simulator}
\subsection{Asymmetric actor-critic framework}
\begin{figure}[!t]
        \centering
        \subfloat[][Teacher policy training.] {
                \includegraphics[width=0.7\linewidth]{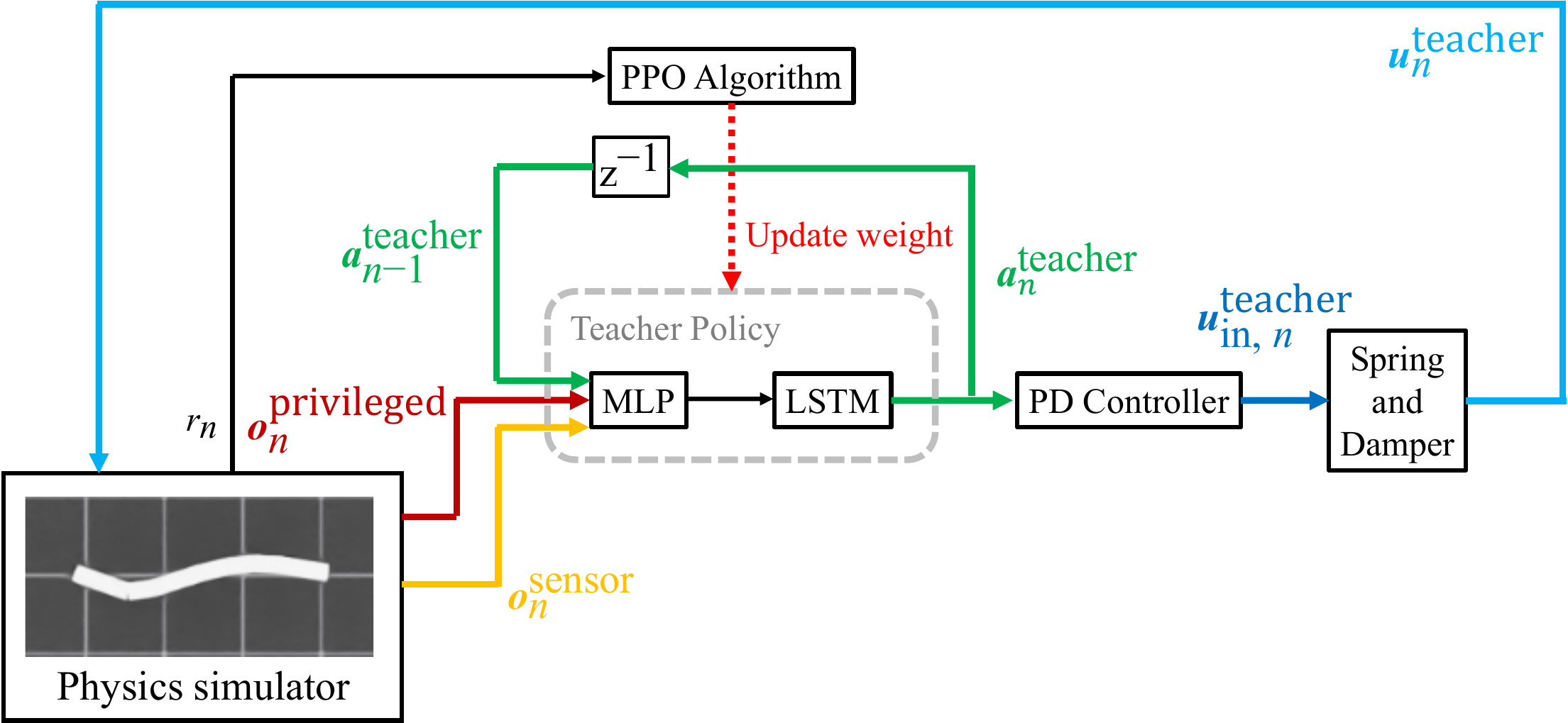}
                \label{fig:teacher_learning}
        }\\
        \subfloat[][Student policy training.] {
                \includegraphics[width=0.7\linewidth]{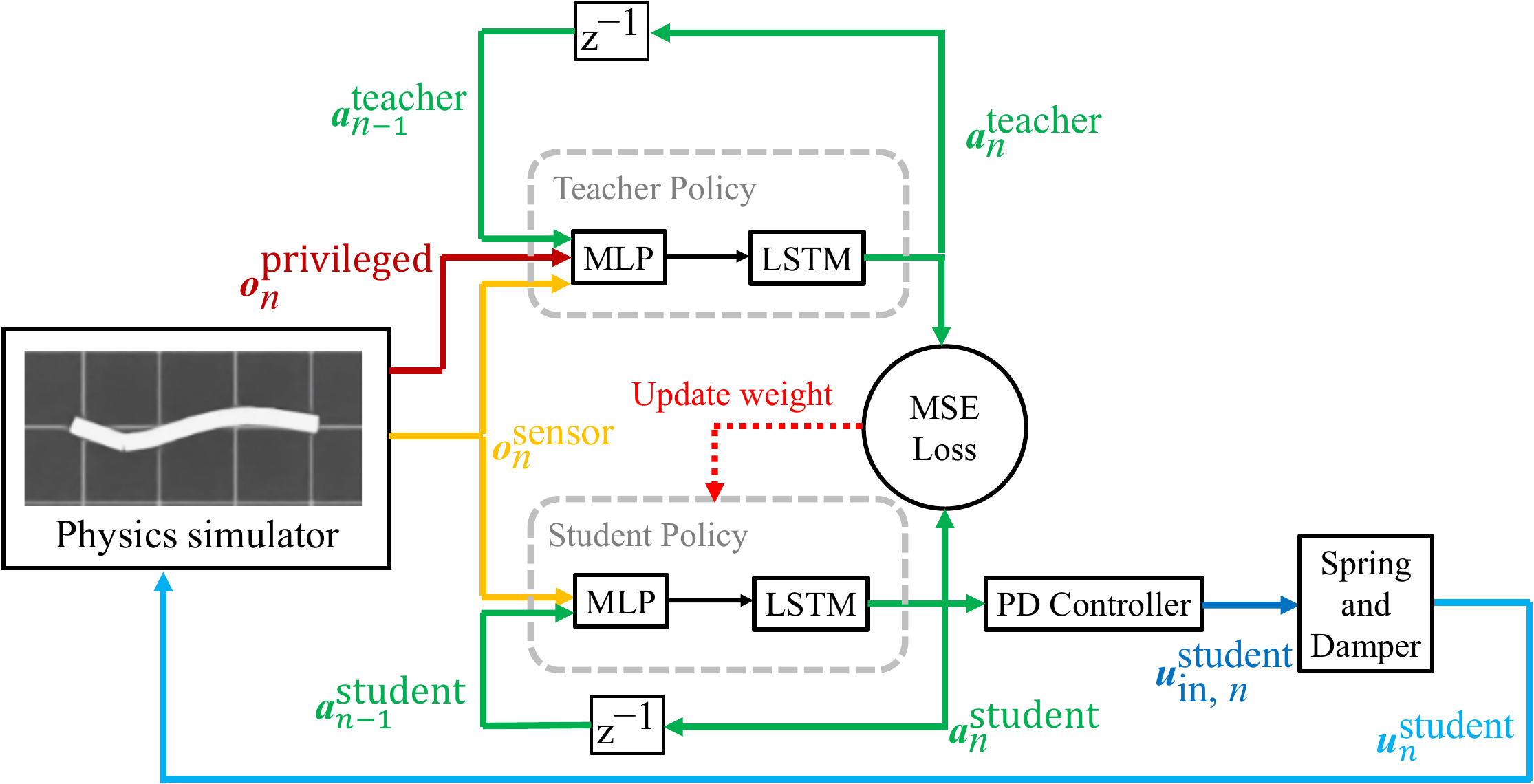}
                \label{fig:student_learning}
        }
        \caption[]{
        Block diagram of the asymmetric actor-critic framework with privileged information. (a) The teacher policy is trained with privileged information that is not available to the student policy. (b) The student policy learns to mimic the teacher's behavior using only sensor information, enabling real-time adaptation to dynamic viscosity changes without external sensors.
        \label{fig:block_diagram}
        }
\end{figure}

To solve the POMDP, we introduce an asymmetric actor-critic framework with privileged information~\cite{Pinto2017}.
The overall structure of this framework is shown in Figure \ref{fig:block_diagram}.
The method consists of a two-stage learning process.
First, a teacher policy $\pi_\mathrm{teacher}$ is trained with privileged information that is only available during training (Figure \ref{fig:teacher_learning}).
Next, the behavior of this teacher policy is distilled into a student policy $\pi_\mathrm{student}$ that relies only on onboard sensor information (Figure \ref{fig:student_learning}).
After acquiring the student policy, the robot can adapt to the environment using only onboard sensor information such as joint angles and torques, without the need to measure or obtain privileged information during real-world deployment.
As a result, real-time gait adaptation to dynamic viscosity changes can be achieved without using external sensors.

\subsection{Observation design}
The observation space of the method consists of two layers: sensor information $\bm{o}_n^\mathrm{sensor}$ and privileged information $\bm{o}_n^\mathrm{privileged}$.
The details of each observation item are summarized in Table~\ref{tab:observations}.
The sensor information includes joint angles, angular velocities, output torques, and the 3-axis angular velocity of the head unit.
In addition, the action from the previous time step is included in the observation space.
The privileged information is only provided during the training of the teacher policy in the simulation.
It includes the translational velocity of the head unit and the kinematic viscosity of the environment.
These physical quantities are difficult to obtain directly from sensors on the actual machine.
The student policy is trained to output optimal target joint angles solely from $\bm{o}_n^\mathrm{sensor}$, excluding the privileged information.

\begin{table}[!t]
\centering
\tbl{Observations and privileged information.}
{\begin{tabular}{llc}
\hline
Category & Variable / Description & Symbol / Dim. \\ \hline
Sensor & Joint angles & $\bm{q}_n \in \mathbb{R}^7$ \\
($\bm{o}_n^\mathrm{sensor}$) & Joint velocities & $\dot{\bm{q}}_n \in \mathbb{R}^7$ \\
& Output torques (Final torque) & $\bm{u}_n \in \mathbb{R}^7$ \\
& Head angular velocity & $\bm{\omega}_n \in \mathbb{R}^3$ \\
& Prev. action (Action at $n-1$) & $\bm{a}_{n-1} \in \mathbb{R}^7$ \\ \hline
Privileged & Fluid kinematic viscosity & $\nu \in \mathbb{R}^1$ \\
($\bm{o}_n^\mathrm{privileged}$) & Head linear velocity & $\bm{v}_n \in \mathbb{R}^3$ \\ \hline
\end{tabular}}
\label{tab:observations}
\end{table}

\subsection{Reward design}
The reward function $r_n$ is designed to balance maximizing propulsion speed and improving energy efficiency.
The total reward is defined as a weighted sum of the following components:
\begin{equation}
r_{n}=w_{1}r_{\mathrm{vel}, n}+w_{2}r_{\mathrm{act}, n}+w_{3}r_{\mathrm{joint\_vel}, n}+w_{4}r_{\mathrm{u\_rate}, n}+w_{5}r_{\mathrm{work}, n}+r_{\mathrm{term}, n}.
\label{eq:reward_function}
\end{equation}
The definitions of each reward component are shown in Table~\ref{tab:reward_definitions}.
The termination penalty $r_{\mathrm{term}, n}$ is of a different from the other reward components.
It is a penalty based on the episode termination condition to ensure the stability of the simulation.
If the robot's center of mass velocity $\|\bm{v}_{n}\|$ exceeds a threshold $v_\mathrm{max}$ at any time step, the episode is terminated and a large negative reward is given.
\begin{equation}
r_{\mathrm{term}, n} \coloneq \begin{cases}
w_6 & \text{if } \|\bm{v}_{n}\| > v_\mathrm{max} \\
0 & \text{otherwise}
\end{cases}.
\end{equation}
Here, $v_\mathrm{max}$ is a velocity threshold for detecting abnormal physical behavior, and $w_6$ is the weight coefficient for the penalty.
This conditional expression helps to suppress the collapse of learning during the exploration phase in the early stages of training.

\begin{table}[t]
\centering
\tbl{Definitions of reward components.}
{\begin{tabular}{lll}
\toprule
Component & Formulation & Description / Goal \\
\midrule
Propulsion & $r_{\mathrm{vel}, n} = v_x$ & Maximizing forward velocity along the $x$-axis \\
Action & $r_{\mathrm{act}, n} = \|\bm{a}_n\|^2$ & Suppressing excessive movements near joint limits \\
Joint velocity & $r_{\mathrm{joint\_vel}, n} = \|\dot{\bm{q}}_n\|^2$ & Promoting smooth undulation by reducing vibrations \\
Torque rate & $r_{\mathrm{u\_rate}, n} = \|\bm{u}_n - \bm{u}_{n-1}\|^2$ & Reducing electrical/mechanical load on servomotors \\
Actuator work & $r_{\mathrm{work}, n} = \sum^{N_\mathrm{servo}}_{i=1} u_{\mathrm{in},n,i} \dot{q}_i$ & Enhancing energy efficiency by minimizing work \\
\bottomrule
\end{tabular}}
\label{tab:reward_definitions}
\end{table}

\section{Experimental setup}
\subsection{Environment setup}
For the training of the DRL policy, we use NVIDIA Isaac Lab (v0.41.1)~\cite{Mittal2025a} for environment setup, and the simulation is conducted on Isaac Sim (v4.5.0) ~\cite{Makoviychuk2021}.
For the state transition calculation, we integrate our original fluid force model into the rigid body dynamics computation of Isaac Sim.
The total fluid force $\bm{f}$ acting on each link is defined as an external force acting on the center of mass.
The total fluid force $\bm{f}$ is a linear combination of the added mass force $\bm{f}_\mathrm{A}$, pressure drag $\bm{f}_\mathrm{F}$, and viscous resistance $\bm{f}_\mathrm{R}$:
\begin{equation}
\bm{f} = \bm{f}_\mathrm{A} + \bm{f}_\mathrm{F} + \bm{f}_\mathrm{R}.
\end{equation}
The added mass force $\bm{f}_\mathrm{A}$ acts in the normal direction of the $i$-th link:
\begin{equation}
\begin{bmatrix}
        f_{\mathrm{A},i}^\mathrm{t} \\ f_{\mathrm{A},i}^\mathrm{n} \end{bmatrix} = -\begin{bmatrix} 0 & 0 \\ 0 & c_{\mathrm{a},i}
\end{bmatrix}
\begin{bmatrix}
        \dot{v}_{\mathrm{t},i} \\ \dot{v}_{\mathrm{n},i}
\end{bmatrix}.
\end{equation}
The pressure drag $\bm{f}_\mathrm{F}$ is proportional to the square of the relative velocity:
\begin{equation}
\begin{bmatrix}
        f_{\mathrm{F},i}^\mathrm{t} \\ f_{\mathrm{F},i}^\mathrm{n} \end{bmatrix}= -
\begin{bmatrix}
        c_{\mathrm{f},i}^\mathrm{t} & 0 \\ 0 & c_{\mathrm{f},i}^\mathrm{n}
\end{bmatrix}
\begin{bmatrix}
        |v_{\mathrm{t},i}|v_{\mathrm{t},i} \\ |v_{\mathrm{n},i}|v_{\mathrm{n},i}
\end{bmatrix}.
\end{equation}
The viscous resistance $\bm{f}_\mathrm{R}$ is calculated with the following formula, which includes the effect of boundary layer thinning \cite{Eloy2013,Yamano2023}:
\begin{equation}
\begin{bmatrix}
        f_{\mathrm{R},i}^\mathrm{t} \\ f_{\mathrm{R},i}^\mathrm{n}
\end{bmatrix}= -
\begin{bmatrix}
        c_{\mathrm{\mu},i}^\mathrm{t} & 0 \\ 0 & c_{\mathrm{\mu},i}^\mathrm{n}
\end{bmatrix}
\begin{bmatrix}
        v_{\mathrm{t},i}\sqrt{|v_{\mathrm{t},i}|} \\ v_{\mathrm{n},i}\sqrt{|v_{\mathrm{n},i}|}
\end{bmatrix}
-
\begin{bmatrix}
        c_{\mathrm{\nu},i}^\mathrm{t} & 0 \\ 0 & 0
\end{bmatrix}
\begin{bmatrix}
        v_{\mathrm{t},i}\sqrt{|v_{\mathrm{n},i}|} \\ 0
\end{bmatrix}.
\end{equation}

The added mass and each drag coefficient are defined as dimensionless quantities using the geometric characteristics of the cylindrical links.
This allows for general applicability that is independent of the type of fluid environment and the physical scale of the robot.
Specifically, using the diameter $d$, length $l$ of each link, fluid density $\rho_\mathrm{f}$, and kinematic viscosity $\nu$, each coefficient with physical dimensions is converted to dimensionless parameters $C_{\mathrm{A},i}, C_{\mathrm{F},i}, C_{\mathrm{D},i}, C_{\mathrm{\mu},i}, C_{\mathrm{\nu},i}$ as follows:
\begin{eqnarray}
  \left.
    \begin{array}{c}
      c_{\mathrm{a}, i} \coloneqq \dfrac{\pi d^2}{4} C_{\mathrm{A}, i}         \rho_\mathrm{f} l, \ c^\mathrm{t}_{\mathrm{f}, i} \coloneqq \dfrac{1}{2} C_{\mathrm{F}, i} \rho_\mathrm{f} \pi d l, \ c^\mathrm{n}_{\mathrm{f}, i} \coloneqq \dfrac{1}{2} C_{\mathrm{D}, i} \rho_\mathrm{f} d l, \\
      c^\mathrm{t}_{\mu, i} \coloneqq C_{\mathrm{\mu}, i} \rho_\mathrm{f} \pi d \sqrt{\nu l}, \ c^\mathrm{n}_{\mu, i} \coloneqq 0, \ c^\mathrm{t}_{\nu, i} \coloneqq C_{\mathrm{\nu}, i} \rho_\mathrm{f} \pi d \sqrt{\nu l}, \ c^\mathrm{n}_{\nu, i} \coloneqq 0
    \end{array}
  \right\}.
\end{eqnarray}
Since the links except for the head and tail have the same shape, the fluid parameters are common to all links, and the parameters are defined as
\begin{eqnarray}
  C_{\mathrm{A}, i} \coloneqq C_\mathrm{A}, \ C_{\mathrm{F}, i} \coloneqq C_\mathrm{F}, \ C_{\mathrm{D}, i} \coloneqq C_\mathrm{D}, \ C_{\mu, i} \coloneqq C_\mu, \ C_{\nu, i} \coloneqq C_\nu.
\end{eqnarray}
The head and tail need to consider the differences in drag characteristics due to their shapes.
Therefore, for the links that have a significant impact on motion, the vertical drag is based on the drag coefficient of the middle links ($i \in \{2, \dots, N-1\}$), $C_\mathrm{D,body}$, and the coefficients for the head ($i=1$) and tail ($i=N$) are defined as constant multiples using the shape constants $p_4, p_5$ for each part:
\begin{eqnarray}
        C_\mathrm{D,head} = p_4 C_\mathrm{D,body}, \quad C_\mathrm{D,tail} = p_5 C_\mathrm{D,body}.
\end{eqnarray}
The drag coefficient \(C_\mathrm{D}\) varies with the Reynolds number
\begin{eqnarray}
  C_\mathrm{D} (\mathrm{Re_n}) \coloneqq 10^{\left\{ p_1 {\left( \log_{10} \mathrm{Re_n} \right)}^2 + p_2 \log_{10} \mathrm{Re_n} + p_3 \right\}},
  \label{eq:cd_approximate_formula}
\end{eqnarray}
where \(\mathrm{Re_n}\) is defined as
\begin{eqnarray}
\mathrm{Re_n} \coloneqq \dfrac{v_\mathrm{n} d}{\nu}.
\end{eqnarray}
This formula approximates the relationship between the Reynolds number and the drag coefficient for a smooth cylinder in steady flow, but it diverges to infinity as \(\mathrm{Re_n} \sim 0\), which contradicts actual phenomena.
To correct this, in the range \(\mathrm{Re_n}<10^{-6}\), the Lamb solution of the Oseen equation for \(\mathrm{Re_n}<10^{-6}\) is referenced, and the equation is smoothly connected to the drag coefficient \cite{Kimoto2025}.
\begin{eqnarray}
C_\mathrm{D} (\mathrm{Re_n}) \coloneqq \dfrac{C_1}{\mathrm{Re_n} \left\{ 1/2 - \gamma - \ln\left({\mathrm{Re_n}/8}\right) \right\}} + C_2,
\end{eqnarray}
Here, the correction parameters \(C_1\) and \(C_2\) ensure a smooth transition with the Euler constant \(\gamma=0.57722\).
The values of the fluid parameters used in the simulation are estimated from experiments and summarized in Table~\ref{tbl:fluid_parameters}~\cite{Yamano2023,Kimoto2025}.

\begin{table}[!t]
        \tbl{Dimensionless parameters for the fluid force model \citep{Yamano2023,Kimoto2025}.}
        {\centering
        \begin{tabular}{cc|c}
        \hline
        & & Values \\
        \hline
        Added mass coefficient & \( C_\mathrm{a} \) & 0.011514 \\
        Tangential inertia resistance coefficient & \( C_\mathrm{f}\) & 0.022556 \\
        Viscous drag coefficient on a link & \( C_{\mu} \) & 0.31584 \\
        Viscous drag coefficient due to boundary layer compression & \( C_{\nu} \) & 1.0612 \\
        \multirow{3}{*}{Normal inertia resistance coefficient \( \left( \mathrm{Re_n} \geq 10^{-6} \right) \)}
        & \( p_1 \) & 0.039276 \\
        & \( p_2 \) & $-0.43683$ \\
        & \( p_3 \) & 0.91538 \\
        \multirow{2}{*}{Normal inertia resistance coefficient \( \left( \mathrm{Re_n} < 10^{-6} \right) \)}
        & \( C_1 \) & 1.36759 \\
        & \( C_2 \) & 2.72646\( \times 10^3 \) \\
        Head scaling factor & \( p_4 \) & 2.9302 \\
        Tail scaling factor & \( p_5 \) & 2.8501 \\
        \hline
        \end{tabular}}
        \label{tbl:fluid_parameters}
\end{table}

\subsection{Training setup}
\subsubsection{DRL algorithm and network structure}
For training the teacher policy, we adopt Proximal Policy Optimization (PPO)~\cite{Schulman2017}.
We use RSL-RL (v2.0.2) as the training library and Adam as the optimization algorithm~\cite{Mittal2025b}.
The network architecture combines a 2-layer MLP (each with 64 units) for feature extraction and a 2-layer LSTM (each with 128 units) for temporal processing.
For the activation function, we use Exponential Linear Units (ELUs), which provide smooth nonlinearity while preventing vanishing gradients.

\subsubsection{Hyperparameters and domain randomization}
The main hyperparameters during training are shown in Table~\ref{tab:hyperparameters}.
In each episode, the kinematic viscosity $\nu$ of the environment is randomized within the range of $\SI{1.0e-7}{m^2/s}$ to $\SI{1.0e-2}{m^2/s}$.
The maximum episode time is set to $\SI{30}{s}$.

\begin{table}[!t]
\centering
\tbl{Hyperparameters for teacher policy training (PPO) and student distillation.}
{
\begin{tabular}{lll}
\hline
Category & Parameter & Value \\ \hline
PPO Algorithm & Learning rate & $1.0 \times 10^{-3}$ \\
 & Clip parameter & $0.2$ \\
 & Discount factor ($\gamma$) & $0.99$ \\
 & GAE parameter ($\lambda$) & $0.95$ \\
 & Num. epochs & $5$ \\
 & Num. mini-batches & $4$ \\ \hline
Distillation & Learning rate & $5.0 \times 10^{-4}$ \\
 & Gradient length & $200$ \\
 & Max iterations & $400$ \\ \hline
Network & MLP hidden dims & $[64, 64]$ \\
 & LSTM hidden dim & $128$ \\
 & LSTM layers & $2$ \\ \hline
\end{tabular}}
\label{tab:hyperparameters}
\end{table}

\subsubsection{Action parameters}
Based on the joint dynamics model established in Section 3.3, we define the specific parameter values used in the simulation.
For the PD controller, the proportional and derivative gains are set to $K_p = 4.0$ and $K_d = 0.3$, respectively, which were determined through trial and error to allow for stable learning.

To ensure a fair comparison between the proposed method and the baseline, the passive physical properties of the joints are set to the same values as in previous research~\cite{Yamano2023}.
The viscous damping coefficient ($c_\mathrm{body} = 0.05 \times 10^{-3} \mathrm{sNm/degree}$) is based on experimental estimates of the effect of the waterproof cover.
On the other hand, the torsional spring constant ($k_\mathrm{body} = 100 \times 10^{-3} \mathrm{ Nm/degree}$) is a value determined in previous research~\cite{Yamano2023} to achieve undulatory locomotion within the specific joint angle and torque constraints of this robot.
In the DRL of this study, the joint angle and torque constraints are the same as those of the snake-like robot, so this constant value is adopted as is. 
During the learning process, the passive restoring force due to this torsional spring plays a role in promoting the agent to acquire undulatory locomotion. 
By storing and releasing elastic energy, it assists the actuators during large-amplitude movements, ensuring that the required active torques safely remain within the physical limits of the hardware for future real-world implementations.

\subsubsection{Reward weights}
The weight coefficients for each term in the reward function are determined based on insights to balance learning stability and propulsion performance (Table \ref{table:reward_weights}).
First, to achieve the primary objective of maximizing propulsion velocity, the main reward $w_1$ is set to a large value.
The weights for $w_3$ (joint velocity) and $w_5$ (work) are kept relatively small to prevent the robot from becoming stationary due to these terms dominating during the early stages of learning when velocities are low.
By fine-tuning $w_2$ (action magnitude) and $w_4$ (torque change rate), we enhance efficiency and stability.
Finally, to avoid overfitting to high-speed movement in low-viscosity environments and the consequent performance degradation in high-viscosity conditions, we gradually decrease $w_1$ to ensure a balanced propulsion performance across the entire viscosity range.

\begin{table}[t]
\centering
\tbl{Weight coefficients for the reward components.}
{\begin{tabular}{lcc}
\toprule
Component & Symbol & Value \\
\midrule
Propulsion velocity & $w_1$ & 10.0 \\
Action magnitude  & $w_2$ & -0.08  \\
Joint velocity       & $w_3$ & -0.002  \\
Torque change rate     & $w_4$  & -0.02  \\
Actuator work          & $w_5$  & -0.005  \\
Termination penalty    & $w_6$  & -10000.0 \\
\bottomrule
\end{tabular}}
\label{table:reward_weights}
\end{table}

\subsection{Evaluation setup}
To quantitatively evaluate the swimming performance of the robot, we use the following four metrics:
\begin{enumerate}
        \item $v$: The average propulsion velocity $v$ is calculated from the movement of the center of mass in the steady state. Specifically, three consecutive peaks are extracted from the velocity time history. Using the distance traveled by the center of mass over a two-cycle interval $(t_1 - t_0)$, it is defined as:
\begin{equation}
v := \frac{w_x(t_1) - w_x(t_0)}{t_1 - t_0},
\end{equation}
where $\bm{w}(t)$ is the coordinate vector of the center of mass, and $w_x(t)$ is its $x$ component.
        \item $P$: As a metric of energy consumption, we use the average power consumption $P$ of all actuators. This is calculated by integrating the product of the commanded torque $u_{\mathrm{in},t, i}$ and the angular velocity $\dot{q}_i$ for each joint over time and averaging it over the cycle:
\begin{equation}
P := \frac{1}{t_1 - t_0} \sum_{i=1}^{N_\mathrm{servo}} \int_{t_0}^{t_1} u_{\mathrm{in},t, i} \dot{q}_i \mathrm{d}t.
\end{equation}
        \item Cost of Transport (CoT): To objectively evaluate the efficiency of movement, we calculate the CoT. This physically represents the amount of energy required to move a unit mass of the robot over a unit distance. Using the previously defined $P$ and $v$, it is defined as:
\begin{equation}
\mathrm{CoT} \coloneqq \frac{P}{mgv},
\end{equation}
where $m$ is the total mass of the robot and $g$ is the gravitational acceleration.
In undulatory locomotion in a fluid, gravitational acceleration does not directly affect the motion, but it is used for non-dimensionalization in the definition of CoT~\cite{Tokic2012,Baines2022}.
        \item $f_\mathrm{swim}$: To analyze the periodicity of the acquired gait, we calculate the propulsion frequency $f_\mathrm{swim}$. We focus on the angle time history of the 7th joint, which has the most stable waveform among all joints. We apply a Fast Fourier Transform (FFT) to this time series data. The most dominant frequency in the resulting spectrum is defined as $f_\mathrm{swim}$.
\end{enumerate}


\section{Results and discussion}
\subsection{Learning convergence and distillation accuracy}
\begin{figure}[!t]
        \centering
        \subfloat[][Comparison of the mean cumulative reward during training with and without privileged information.] {
                \includegraphics[width=0.7\linewidth]{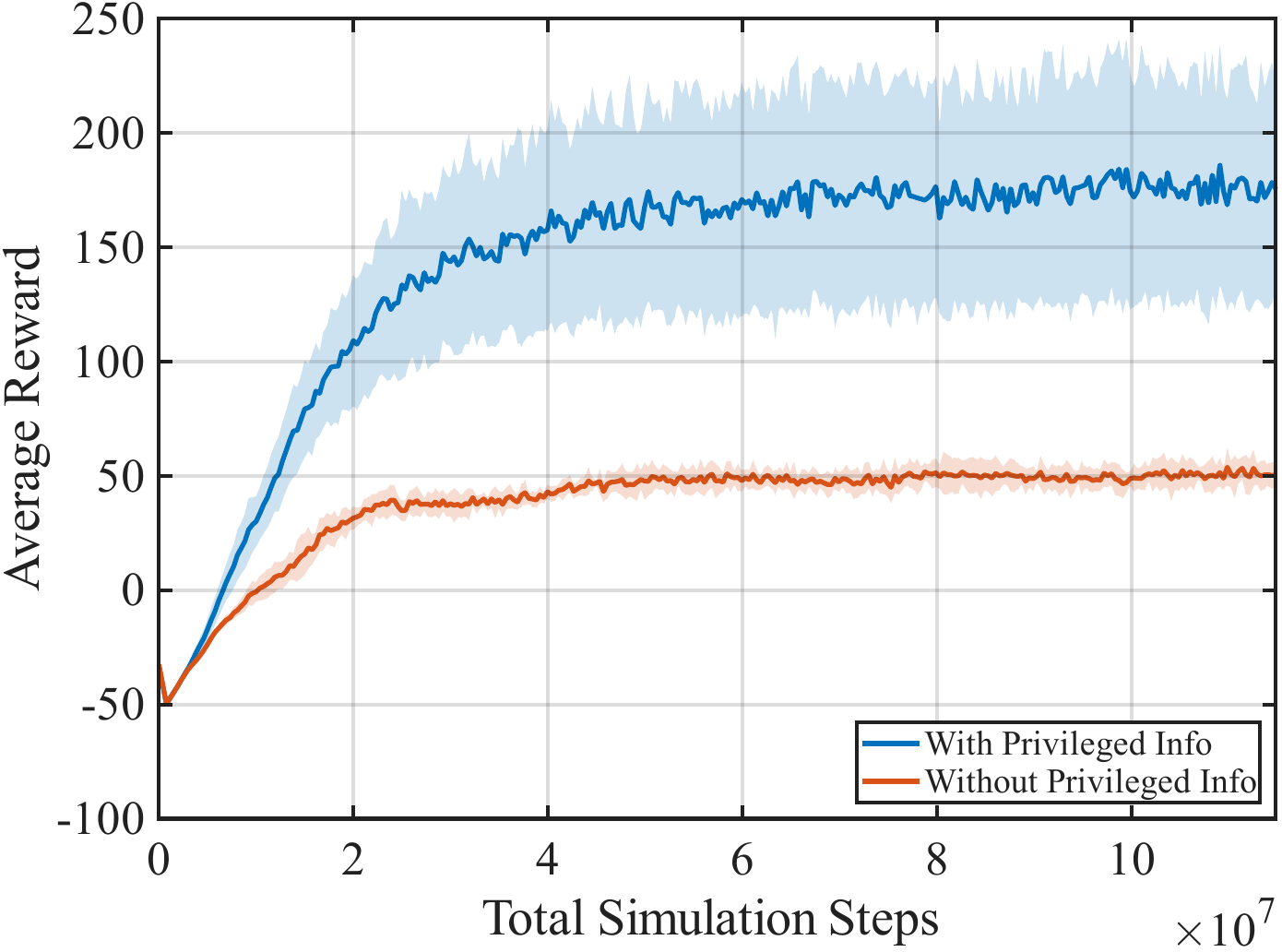}
                \label{fig:reward_curve}
        }\\
        \subfloat[][MSE loss curve during the distillation process.] {
                \includegraphics[width=0.7\linewidth]{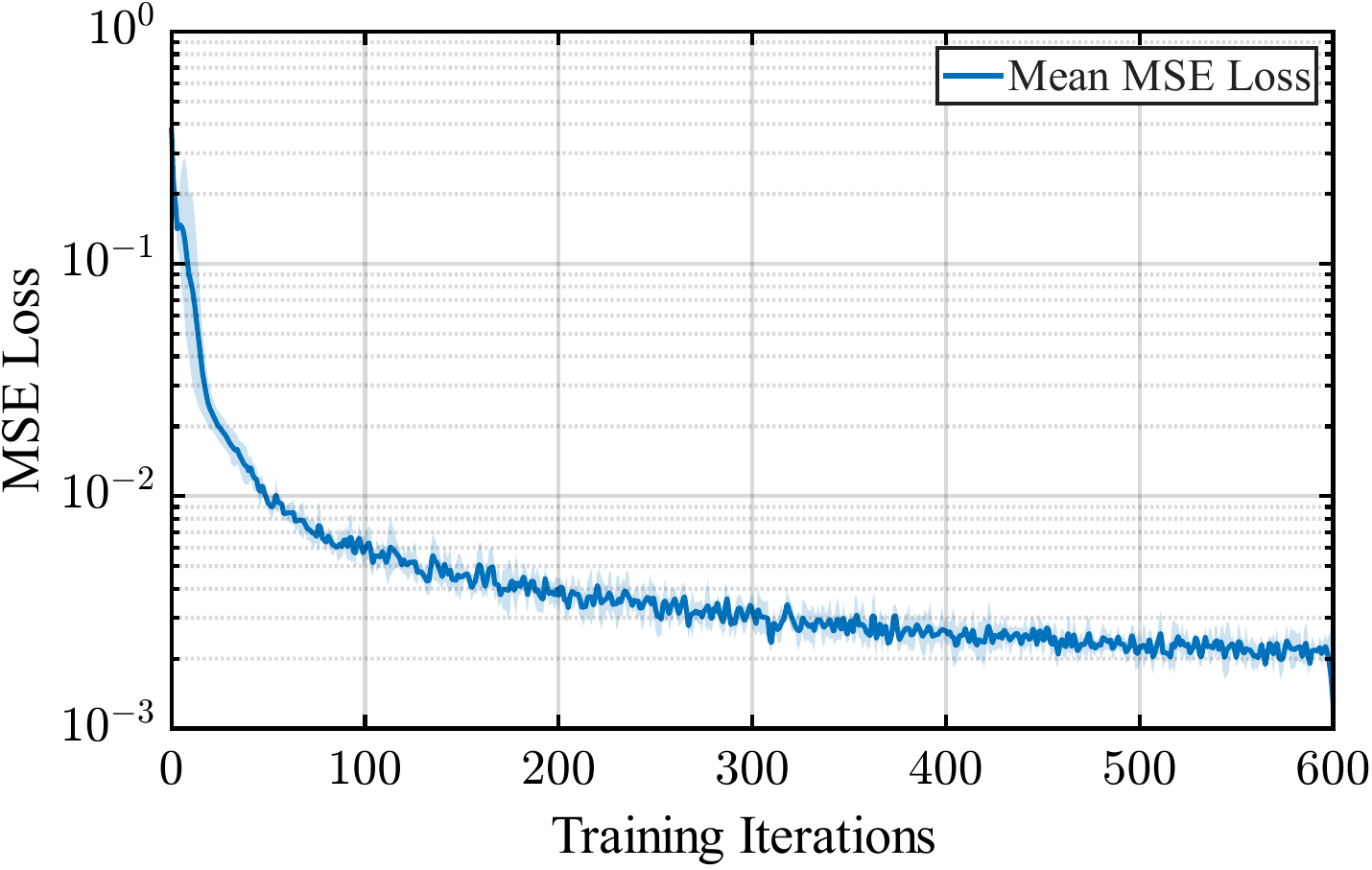}
                \label{fig:distillation_loss_curve}
        }
        \caption[]{
        Training and distillation performance of the proposed framework. (a) Comparison of the mean cumulative reward during training. The proposed method (with privileged information) successfully converges to a high reward, whereas the standard symmetric reinforcement learning approach (without privileged information) plateaus at a lower value, demonstrating the necessity of the asymmetric actor-critic architecture for this task. (b) Mean squared error (MSE) loss during the knowledge distillation process from the teacher to the student policy. The loss consistently decreases, indicating successful behavior cloning using only sensor observations.
        \label{fig:learning_results}
        }
\end{figure}

The training of the teacher policy was conducted under the environment with randomized kinematic viscosity \( \nu \) in the range from \( 10^{-7} \) to \( 10^{-2} \mathrm{m}^2/\mathrm{s} \).
As shown in Figure~\ref{fig:reward_curve}, the mean cumulative reward steadily improved as training progressed and eventually reached a nearly constant value.
This stable convergence indicates that a single policy has acquired a robust propulsion strategy across a wide range of fluid environments.
To ensure the statistical validity of the results, the training and evaluation were conducted over five independent trials with different random seeds, and the curves and shaded areas in the figures represent the mean and standard deviation, respectively.

To clearly demonstrate the necessity of the proposed asymmetric actor-critic structure in this task, Figure~\ref{fig:reward_curve} also shows the learning curve of a standard symmetric DRL approach trained without privileged information (relying solely on sensor information).
The proposed method converges with an average reward of around 170, whereas this standard approach plateaus at an average reward of around 50, failing to acquire an effective swimming gait.
This is because, in the absence of sensors that directly measure changes in fluid viscosity, the dynamic changes in the environment become unobservable for a standard agent.
This comparison result demonstrates that privileged information is critical for locomotion performance.
Consequently, formulating this task as a POMDP to implicitly estimate these unobservable environmental characteristics through knowledge distillation is an essential approach to achieve adaptive swimming in unknown viscous environments.

Next, we describe the results of the knowledge distillation from the teacher policy to the student policy.
Figure~\ref{fig:distillation_loss_curve} shows the transition of the mean squared error between the teacher and student actions during the distillation process.
The MSE decreased from the beginning of training and reached a nearly constant value with a small decrease around 600 iterations.
This result suggests that the student policy can accurately mimic the teacher's decision-making solely from sensor information \( \bm{o}_t^\mathrm{sensor} \) and action history \( \bm{a}_{t-1} \) without access to privileged information.

\subsection{Comparison with Pareto optimal solutions}
To evaluate the performance of the proposed method, we compared it with two traditional control frameworks, namely SOO and MOO, as described in Section 3.4.
The SOO is based on a kinematic sea snake model optimized with the same objective function as the DRL, while the MOO provides the strict performance limits (Pareto fronts) of predefined sinusoidal torque control.
In this evaluation, to maintain consistency with previous research~\cite{Yamano2023}, we selected water and two types of oil with different kinematic viscosities (Oil I, Oil II) as representative evaluation environments.
The specific physical properties of these fluid environments are summarized in Table~\ref{tbl:fluid_parameters}.
The comparison was conducted from two perspectives: specialization performance for specific environments and generalization performance across all environments.

\begin{figure}[!t]
        \centering
        \subfloat[][Performance comparison with environment-specific Pareto fronts for water, Oil I, and Oil II.] {
                \includegraphics[width=0.6\linewidth]{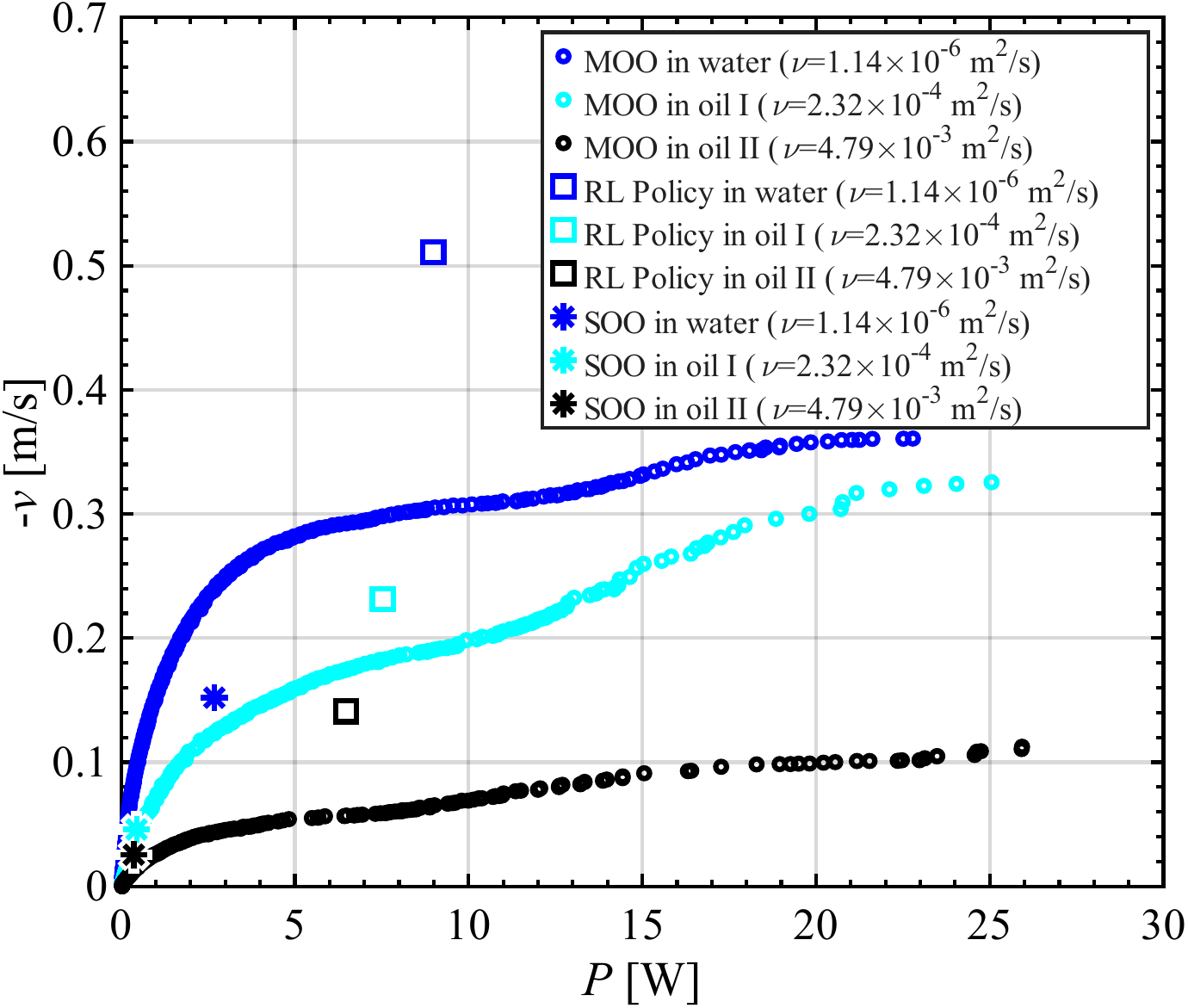}
                \label{fig:pareto_vs_learning}
        }\\
        \subfloat[][Comparison of the global performance distribution.] {
                \includegraphics[width=0.6\linewidth]{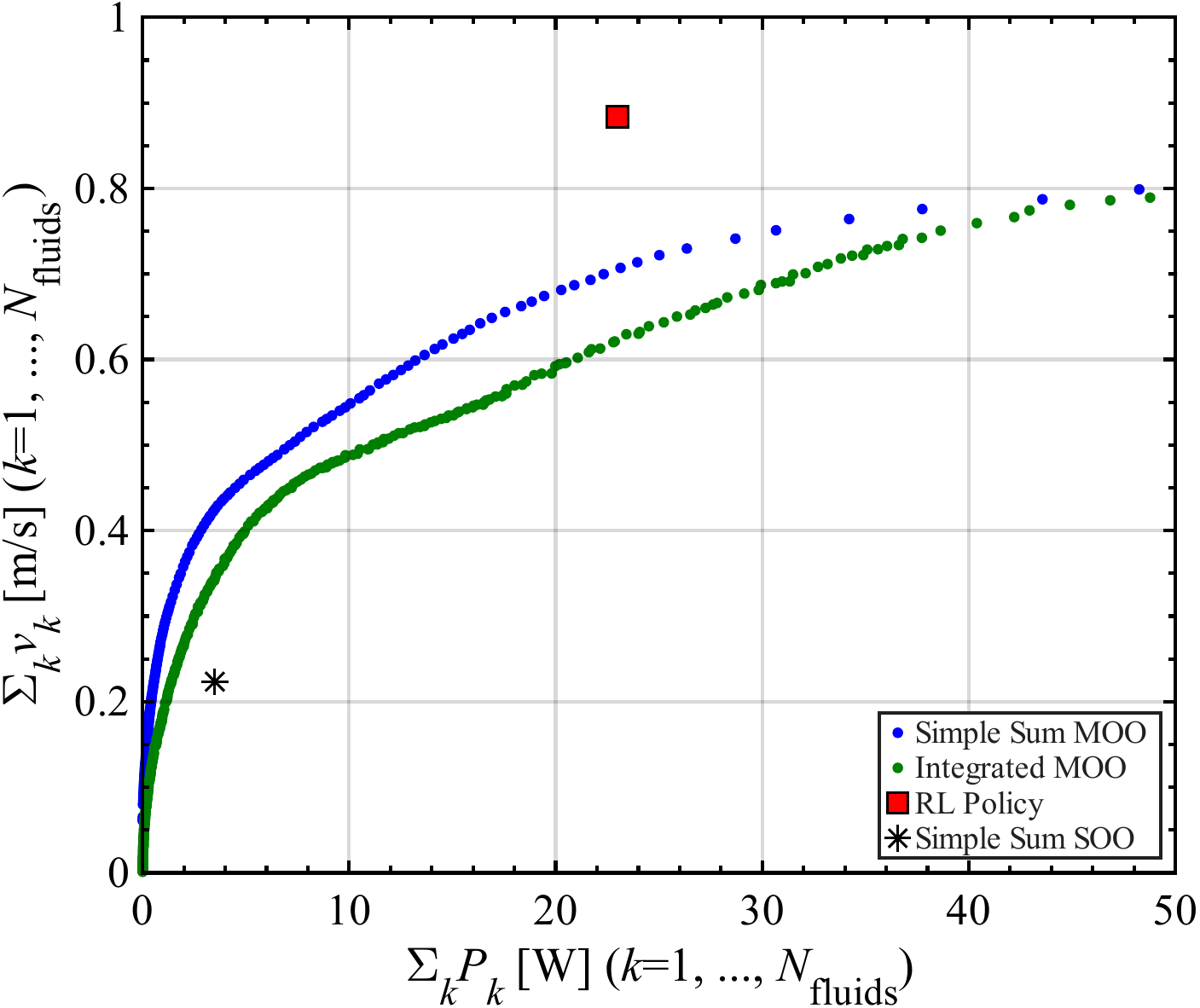}
                \label{fig:integrated_pareto_front_vs_learning}
        }
        \caption[]{
        Comparison of the proposed method with Pareto optimal solutions obtained by MOO based on sinusoidal torque control and sea snake kinematic SOO. (a) Performance comparison with environment-specific Pareto fronts for water, Oil I, and Oil II. The proposed method achieves higher propulsion velocity than both the Pareto fronts and the sea snake kinematic SOO across all environments. (b) Comparison of the global performance distribution. The proposed method surpasses both the integrated MOO and simple sum MOO, indicating that it autonomously acquires dynamic non-sinusoidal gaits that exceed the limitations of sinusoidal control frameworks. $N_\mathrm{fluids} = 3$ is the number of fluid environments.
        \label{fig:pareto_comparison}
        }
\end{figure}

Figure~\ref{fig:pareto_vs_learning} compares the performance of the Pareto fronts individually optimized for each environment, the proposed method, and sea snake kinematic SOO.
In all environments, the proposed method achieved propulsion performance that surpasses the environment-specific Pareto fronts.
Furthermore, the comparison with sea snake kinematic SOO (star markers) reveals that while the traditional kinematic model (SOO) tends to sacrifice propulsion speed in favor of energy efficiency, the proposed method is freed from the efficiency-focused constraints of SOO and autonomously acquires non-sinusoidal waveforms that achieve higher-speed propulsion.
This suggests that the proposed method goes beyond the framework of sinusoidal control and generates dynamic motions that actively overcome fluid resistance.

To evaluate the generalization performance against unknown environmental changes, we compared the global performance distribution in Figure~\ref{fig:integrated_pareto_front_vs_learning}.
\begin{itemize}
        \item Integrated MOO (green): This represents the limit of using a single fixed sinusoidal parameter across all kinematic viscosity environments, indicating the performance of a fixed control strategy that does not account for environmental changes.
        \item Simple Sum MOO (blue): This is obtained by integrating the Pareto fronts obtained individually for each environment and extracting the Pareto optimal solutions again. This represents the ideal adaptive performance if the optimal parameters could be instantaneously switched according to the environment while still assuming sinusoidal control.
\end{itemize}
The proposed method achieved high propulsion performance and efficiency that even surpassed the Simple Sum MOO.
This indicates that the proposed method is dynamically generating joint waveforms that are not constrained by the framework of sinusoidal control, tailored to the characteristics of fluid resistance.
It achieved performance that exceeds the ideal adaptive sinusoidal control with a single network under unknown viscosity environments.
We set the number of joints to 7 for consistency with previous research~\cite{Yamano2023}, but since increasing the number of joints allows for approximating smoother curves, there is a potential for further performance improvement with this method.

\begin{table}[t]
  \centering
  \tbl{Physical properties of the fluids used in the simulation \cite{Yamano2023,Kimoto2025}.}
  {\setlength{\tabcolsep}{4pt} 
  \begin{tabular}{lccc} 
    \toprule
    Fluid & Water & Oil I & Oil II \\
    \midrule
    \makecell[l]{Kinematic viscosity \\ $\nu$ [mm$^2$/s]} \\
    \quad @ 15$^{\circ}$C & 1.14 & 232 & $4.79 \times 10^3$ \\
    \quad @ 40$^{\circ}$C & 0.658 & 55.7 & 689 \\
    \quad @ 100$^{\circ}$C & 0.294 & 7.77 & 43.2 \\
    \midrule
    Density $\rho_\mathrm{f}$ [kg/m$^3$] (15$^{\circ}$C) & 999 & 874 & 900 \\
    \midrule
    Product name & \makecell[c]{Tap \\ water} & \makecell[c]{SUPER \\ HYRANDO 56} & \makecell[c]{BONNOC \\ TS 680} \\
    \bottomrule
  \end{tabular}}
   \label{tbl:specifications_of_viscous_fluids}
\end{table}

\subsection{Analysis of generated gaits and verification of the effectiveness of adaptation}
\subsubsection{Adaptation of spatial and temporal gait characteristics}
\begin{figure*}[!t]
        \centering
        \includegraphics[width=\linewidth]{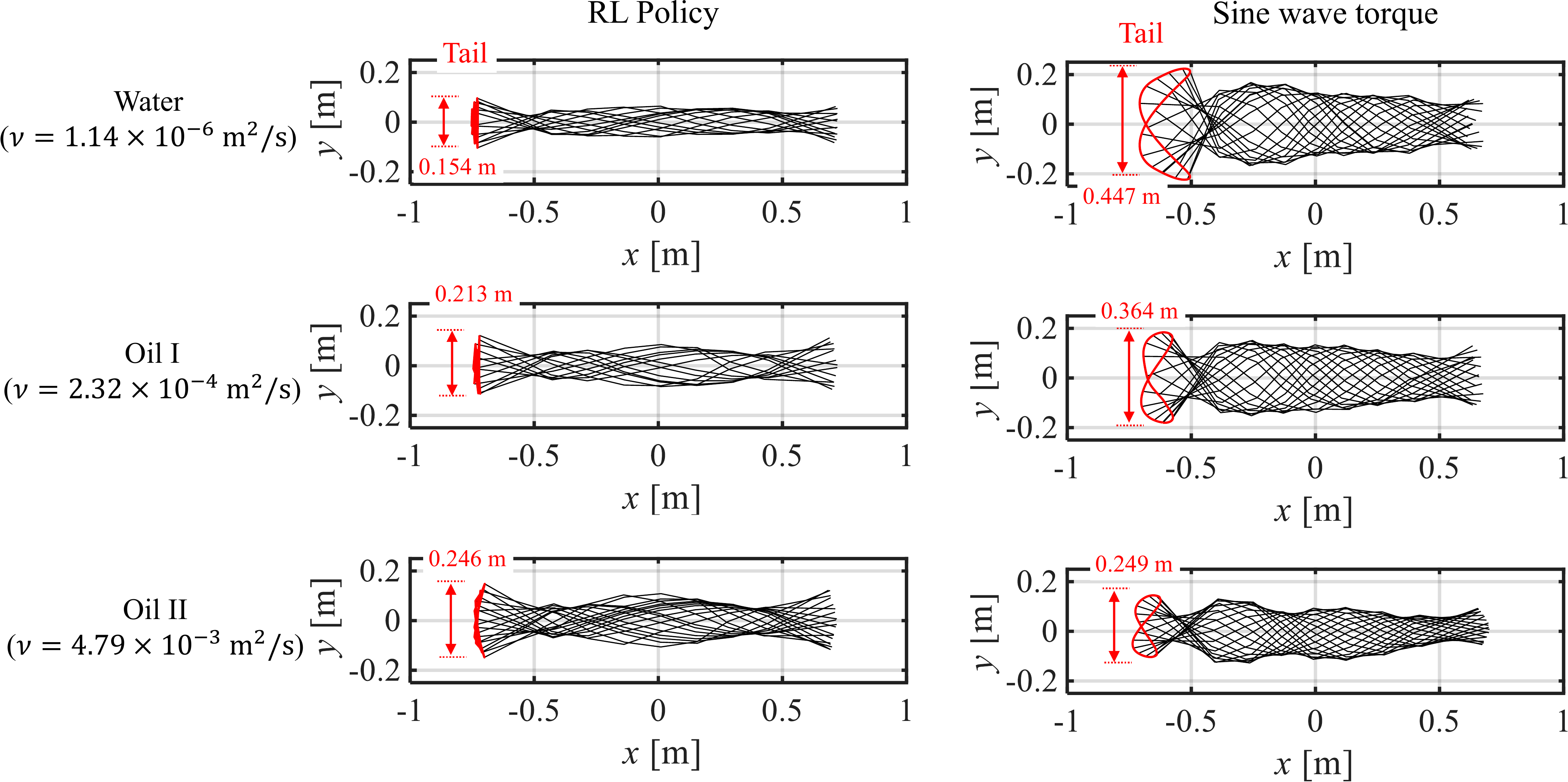}
        \caption[]{
Snapshots of robot poses during steady swimming in different kinematic viscosity environments: (a) Water ($\nu = \SI{1.14e-6}{\meter^2\per\second}$), (b) Oil I ($\nu = \SI{2.32e-4}{\meter^2\per\second}$), and (c) Oil II ($\nu = \SI{4.79e-3}{\meter^2\per\second}$). The left column shows the snapshots generated by the proposed DRL policy, while the right column shows those generated by a specific sinusoidal torque control ($A_\mathrm{in} = \SI{3.5}{\newton\meter}$, $f_\mathrm{cmd} = \SI{0.70}{\hertz}$, $\phi_\mathrm{in} = \SI{-0.84}{\radian}$).
        }
        \label{fig:snapshots}
\end{figure*}

Figure~\ref{fig:snapshots} compares the snapshots of the steady swimming generated by the proposed DRL policy (left) and a specific sinusoidal torque control (right) under each viscosity environment.
In the sinusoidal control on the right side of Figure~\ref{fig:snapshots}, there is a tendency for the amplitude of the tail to decrease with increasing kinematic viscosity (for example, the tail amplitude decreases from 0.447 m in water to 0.249 m in Oil II).
This is because as the kinematic viscosity increases, the resistance increases, and when a certain torque waveform ($A_\mathrm{in} = \SI{3.5}{\newton\meter}$, $f_\mathrm{cmd} = \SI{0.70}{\hertz}$, $\phi_\mathrm{in} = \SI{-0.84}{\radian}$) is input, the displacement at the joints is suppressed.

In contrast, in the proposed DRL policy on the left side of Figure~\ref{fig:snapshots}, there is a tendency for the amplitude of the tail to increase with increasing kinematic viscosity (for example, the tail amplitude increases from 0.154 m in water to 0.246 m in Oil II).
This result shows a similar trend to that exhibited by organisms performing undulatory locomotion~\cite{Stin2024}.
While the reason for this is not fundamentally clear, it has been pointed out that the amplitude of the tail, in particular, affects the stride length when velocity is non-dimensionalized~\cite{Kimoto2025}.
It is considered that the tail amplitude was increased to ensure sufficient thrust against the increasing resistance.

\begin{figure}[!t]
        \centering
        \subfloat[][Time-series of joint torques in water.] {
                \includegraphics[width=0.7\linewidth]{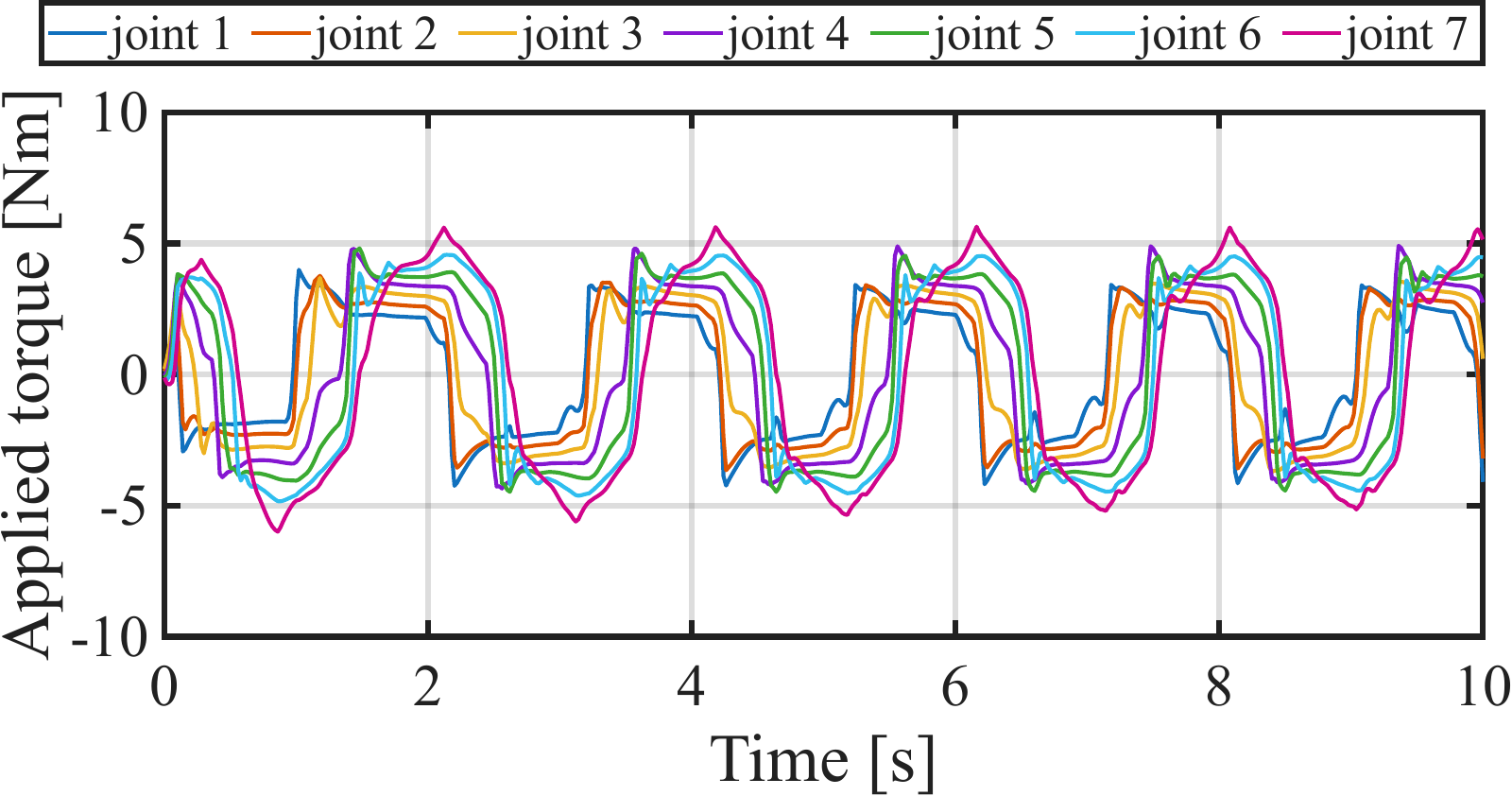}
                \label{fig:applied_torque_water}
        }\\
        \subfloat[][Time-series of joint torques in Oil I.] {
                \includegraphics[width=0.7\linewidth]{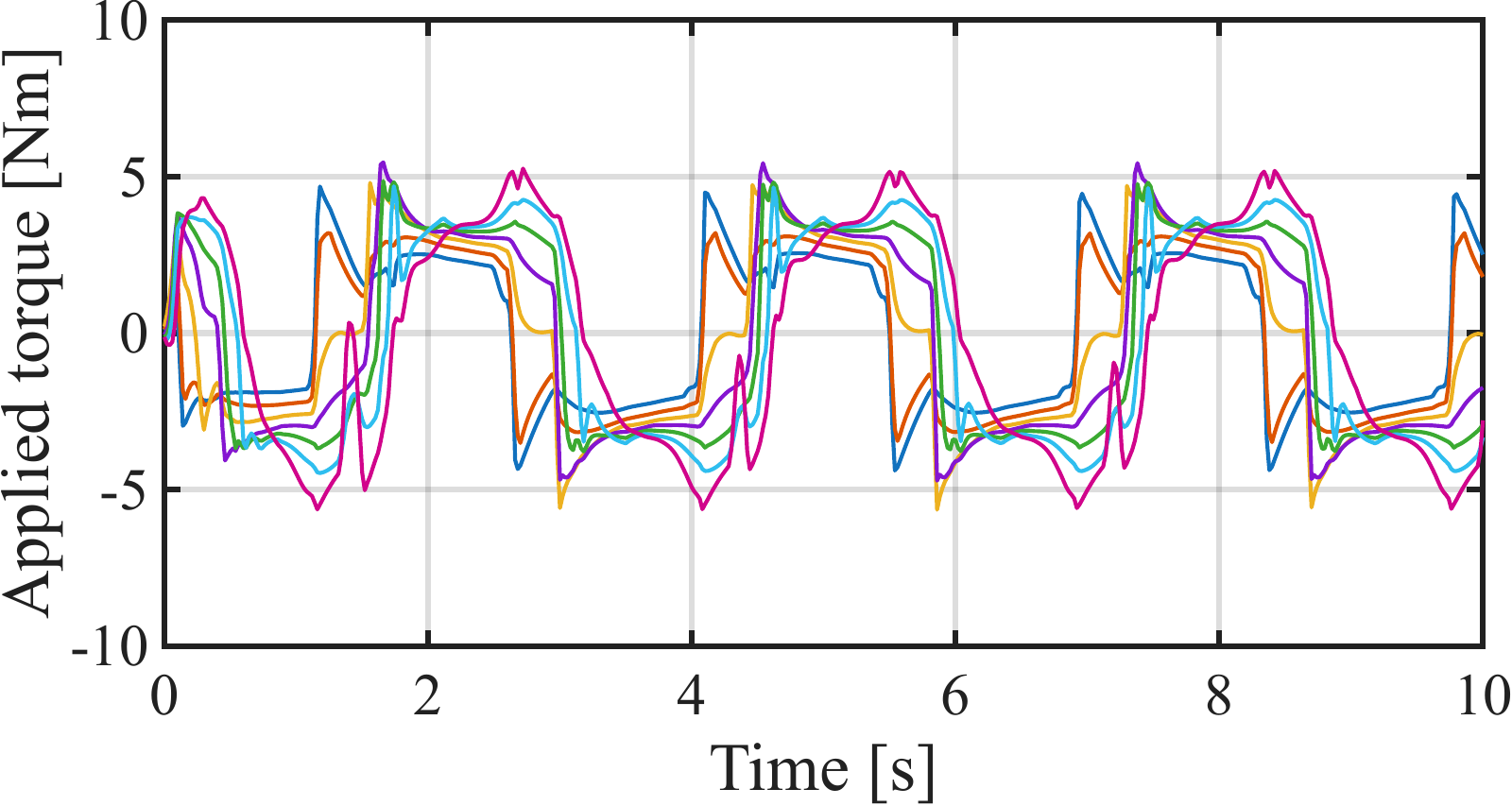}
                \label{fig:applied_torque_oilI}
        }\\
        \subfloat[][Time-series of joint torques in Oil II.] {
                \includegraphics[width=0.7\linewidth]{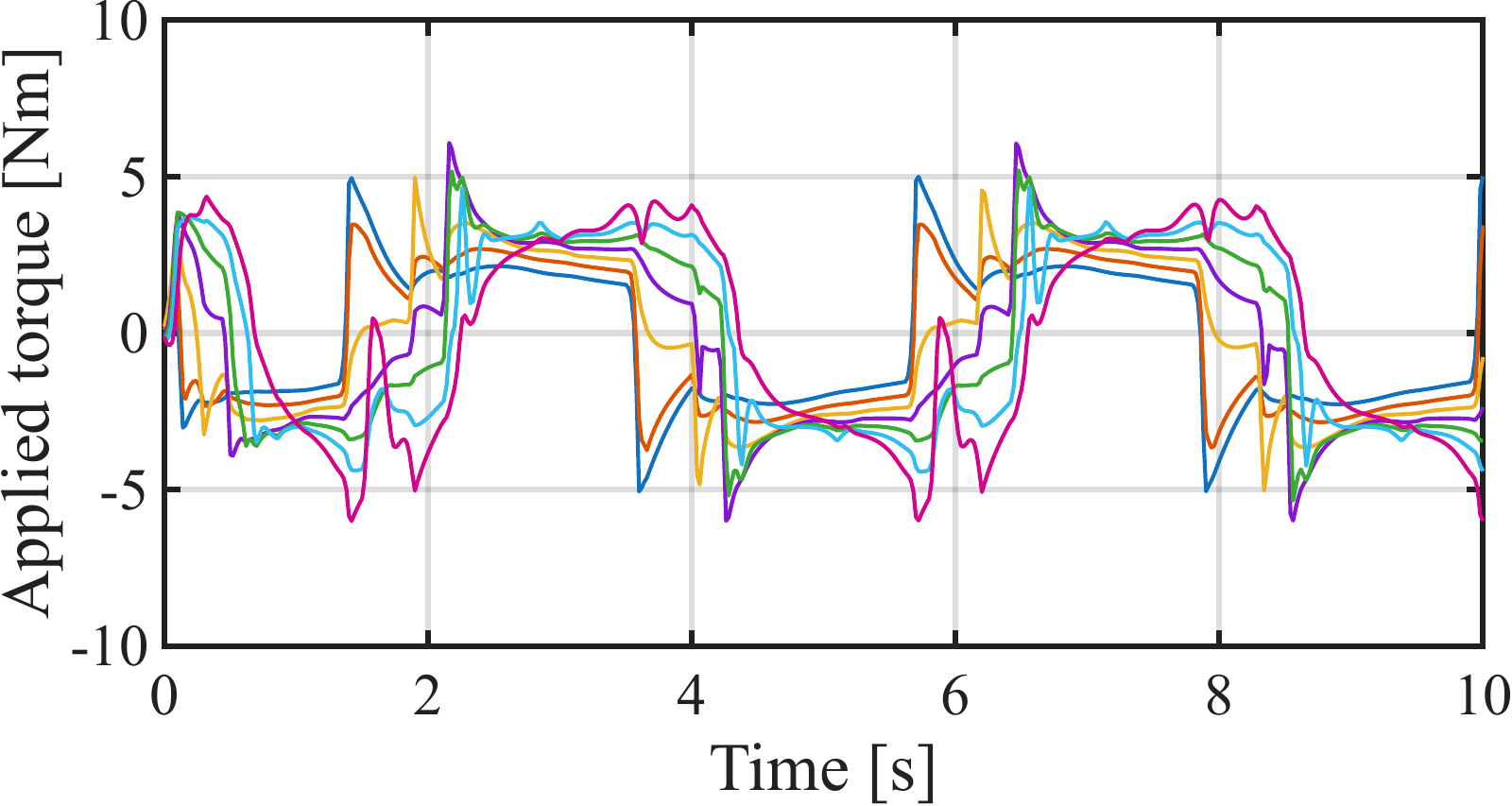}
                \label{fig:applied_torque_oilII}
        }
        \caption[]{
        Time-series of joint torques $u_n$ in different kinematic viscosities: (a) Water ($\nu = \SI{1.14e-6}{\meter^2\per\second}$), (b) Oil I ($\nu = \SI{2.32e-4}{\meter^2\per\second}$), and (c) Oil II ($\nu = \SI{4.79e-3}{\meter^2\per\second}$). The undulation frequency adaptively decreases as the viscosity increases, resulting in a longer period for each cycle.
        \label{fig:applied_torque_time_series}
        }
\end{figure}

Next, we examine the temporal and dynamic characteristics of the gait.
From Figure~\ref{fig:applied_torque_time_series}, it can be observed that the frequency of the undulatory locomotion gradually decreases with increasing kinematic viscosity.
Specifically, while the frequency was about 0.47 Hz in the water environment, it decreased to about 0.23 Hz in the Oil II environment.
This change is similar to the behavior exhibited by organisms performing undulatory locomotion~\cite{Horner2008,Iwasaki2014,Gjorgjieva2014}.

While the frequency decreases with increasing kinematic viscosity, the amplitude of the final output torque $u_n$ is maintained almost constant in the range of 4 to 5 Nm. 
It should be noted that the torque time-series data presented here represent the actual exerted torque integrating the passive elastic forces (as formulated in Eq.~\eqref{eq:torque_generation}, rather than the command torque $u_{\mathrm{in},n}$.
One possible factor for the increase in tail displacement without an increase in torque amplitude is the suppression of fluid resistance due to the decrease in frequency.
Since viscous resistance from the fluid depends on velocity (frequency), lowering the frequency can reduce the instantaneous resistance force experienced by the links~\cite{Kimoto2025}.
This allows for deeper bending of the joints against high viscous resistance even within the limited range of torque output.
This transition to low frequency, large tail amplitude gaits is a result of the DRL autonomously acquiring a strategy to ensure sufficient thrust under high viscosity while suppressing energy consumption.

\subsubsection{Necessity for Adaptation via Cross-Environment Evaluation}
We conducted a cross-environment evaluation where the steady-state torque waveforms generated by the proposed DRL policy for specific environments as shown in Figure \ref{fig:applied_torque_time_series} were directly applied as fixed commands to different viscosity environments.
This allows us to verify the effectiveness of the policy adapting and changing the gait according to the environment.
The evaluation metrics used were propulsion velocity $v$ and CoT.
From the results in Table~\ref{tab:cross_eval}, it can be confirmed that in all fluids, the gait adapted to that fluid (diagonal elements) simultaneously achieves the highest propulsion velocity and the lowest CoT.
In the water environment, the adapted gait (Water-gait) recorded the highest velocity of 0.51 m/s and the lowest CoT of 0.33.
In contrast, when applying the Oil II-gait, which is adapted for high viscosity, to the water environment, the velocity decreased to 0.15 m/s and the CoT also deteriorated to 0.42.
This indicates that using an excessively low frequency under low viscosity is disadvantageous in terms of both propulsion performance and efficiency.

In the Oil II environment, the adapted gait (Oil II-gait) achieved a velocity of 0.14 m/s and a CoT of 0.85, showing superior performance compared to non-adapted gaits.
In particular, when applying the Water-gait, which is adapted for water, to the Oil II environment, the velocity drastically decreases to 0.06 m/s and the CoT deteriorates significantly to 1.62.
In high viscosity, selecting an appropriate frequency and large amplitude that matches the resistance characteristics is a decisive factor in achieving efficient propulsion.

\begin{table}[!t]
\centering
\tbl{Performance matrix of cross-environment evaluation.}
{\begin{tabular}{lcccccc}
\toprule
Test Environment & \multicolumn{2}{c}{Water-gait} & \multicolumn{2}{c}{Oil I-gait} & \multicolumn{2}{c}{Oil II-gait} \\
\midrule
& $v$ & $\mathrm{CoT}$ & $v$ & $\mathrm{CoT}$ & $v$ & $\mathrm{CoT}$ \\
Water ($\nu = \SI{1.14e-6}{\meter^2\per\second}$) & \textbf{\SI[detect-weight]{0.51}{m/s}} & \textbf{0.33} & \SI{0.22}{m/s} & 0.39 & \SI{0.15}{m/s} & 0.42 \\
Oil I ($\nu = \SI{2.32e-4}{\meter^2\per\second}$) & \SI{0.12}{m/s} & \SI{0.87}{} & \textbf{\SI[detect-weight]{0.23}{m/s}} & \textbf{0.60} & \SI{0.09}{m/s} & 0.73 \\
Oil II ($\nu = \SI{4.79e-3}{\meter^2\per\second}$) & \SI{0.06}{m/s} & 1.62 & \SI{0.06}{m/s} & 1.38 & \textbf{\SI[detect-weight]{0.14}{m/s}} & \textbf{0.85} \\
\bottomrule
\end{tabular}}
\label{tab:cross_eval}
\end{table}

\section{Conclusion}
In this study, we demonstrated how DRL can enable snake-like robots to autonomously perform adaptive swimming in unknown fluid environments with dynamically changing viscosity, effectively overcoming the inherent performance limitations of traditional control methods.
A key approach was to treat the dynamic viscosity and the translational velocity of the head link as unobservable hidden variables in the swimming phenomenon of the snake-like robot, and to formulate the problem as a POMDP.
By applying the framework of asymmetric actor-critic that leverages privileged information in the physics simulator, we successfully adapted to viscosity changes.
Through distillation of the knowledge from the teacher policy to the student policy, we established a control system that generates optimal joint angles in real-time solely from onboard sensor information without any external sensors.
The necessity of this framework was validated by demonstrating that a standard symmetric DRL approach fails to acquire an effective gait in this partially observable setting.

The results of the evaluation through physics simulation showed that the proposed method achieved propulsion velocity and cost of transport that even surpassed the ideal adaptive (Simple Sum MOO) based on traditional sinusoidal control.
Through gait analysis, it was revealed that the robot autonomously acquired a physically reasonable strategy of lowering the frequency while increasing the tail amplitude under high-viscosity environments.
These findings demonstrate that the proposed method overcomes the inherent performance limitations of traditional control frameworks in dynamic viscous environments.

Having established this fundamental advantage, we plan to investigate Sim-to-Real transfer for applying this method to real hardware in the future.
Because the proposed framework distills knowledge into a student policy relying solely on onboard proprioceptive sensors, it is structurally well-suited for physical deployment without requiring impractical external fluid measurement devices.
To successfully bridge the Sim-to-Real gap, our next step is to address hardware-specific challenges that are difficult to capture in simulation, such as the non-linear dynamics of the waterproof cover and the response delays of the actual actuators under high fluid resistance.
Furthermore, we will verify the impact of increasing or decreasing the number of joints on performance, and further improve adaptability to complex heterogeneous environments.

\section*{Acknowledgements}
The authors would like to thank Gemini 3 Flash (Google) for its assistance in English proofreading and language improvement. All AI-generated suggestions were critically reviewed and verified by the authors, who remain solely responsible for the final content and scientific accuracy of this work.

\section*{Disclosure statement}

No potential conflict of interest was reported by the author(s).

\section*{Funding}

This study was supported by a Grant-in-Aid for Scientific Research (C) (No.22K03983) and JST SPRING, Grant Number JPMJSP2139.

\section*{Notes on contributors}
\emph{Tsuyoshi Kimoto} received B. Eng and M. Eng in Aerospace Engineering from Osaka Prefecture University, Sakai, Japan in 2020 and 2022, respectively. Since 2022, he is a Doctoral student in Osaka Metropolitan University, Sakai, Japan. His research interests include flexible robots and autonomous systems.

\vskip\baselineskip
\noindent
\emph{Akio Yamano} received B.Eng., M.Eng., and Ph.D. degrees in mechanical engineering from Osaka Prefecture University, Osaka, Japan, in 2012, 2014, and 2017, respectively. Since 2025, he has been a Lecturer at the Department of Aerospace Engineering, Graduate School of Engineering, Osaka Metropolitan University. His research interests include fluid–structure interaction, design of flexible robots, and self-excited oscillators.

\vskip\baselineskip
\noindent
\emph{Kohei Honda} received his Ph.D. in Mechanical Systems Engineering from the Graduate School of Engineering, Nagoya University, in 2024. In the same year, he was a visiting researcher at the University of California, Berkeley. He is currently an Assistant Professor at the Graduate School of Engineering, Nagoya University, and a Research Scientist at CyberAgent AI Lab under a cross-appointment. His research interests include motion planning based on probabilistic inference and optimal control theory, as well as the integration of model-based methods and machine learning for intelligent autonomous mobile robots.

\vskip\baselineskip
\noindent
\emph{Takashi Iwasa} received D.Eng. in Aerospace Engineering from University of Tokyo, Japan in 2004. Since 2022, he has been a Professor of aerospace engineering at the Osaka Metropolitan University. His research interests include mechanics of membrane space structures, photogrammetric measurement for space structures, and shock test for space equipment.

\bibliographystyle{tfnlm}
\bibliography{interactnlmsample}

\end{document}